# WORD SENSE DISAMBIGUATION USING WSD SPECIFIC WORDNET OF POLYSEMY WORDS

Udaya Raj Dhungana[1], Subarna Shakya[2], Kabita Baral[3] and Bharat Sharma[4]

[1, 2, 4]Department of Electronics and Computer Engineering, Central Campus, IOE, Tribhuvan University, Lalitpur, Nepal
[1]udayas.epost@gmail.com
[2]drss@ioe.edu.np
[4]vharatsharma@yahoo.com
[3]Department of Computer Science, GBS, Lamachaur, Kaski, Nepal
[3]kabitabaral@yahoo.com

## ABSTRACT

*This paper presents a new model of WordNet that is used to disambiguate the correct sense of polysemy word based on the clue words. The related words for each sense of a polysemy word as well as single sense word are referred to as the clue words. The conventional WordNet organizes nouns, verbs, adjectives and adverbs together into sets of synonyms called synsets each expressing a different concept. In contrast to the structure of WordNet, we developed a new model of WordNet that organizes the different senses of polysemy words as well as the single sense words based on the clue words. These clue words for each sense of a polysemy word as well as for single sense word are used to disambiguate the correct meaning of the polysemy word in the given context using knowledge based Word Sense Disambiguation (WSD) algorithms. The clue word can be a noun, verb, adjective or adverb.*

## KEYWORDS

*Word Sense Disambiguation, WordNet, Polysemy Words, Synset, Hypernymy, Context word, Clue Words*

## 1. INTRODUCTION

Words that express two or more different meanings when used in different contexts are referred to as polysemy or multi-sense words. Every natural language contains such polysemy words in its vocabulary. Such polysemy words create a big problem during the translation of one natural language to another. To translate the correct meaning of the polysemy word, the machine must first know the context in which the polysemy word has been used. Only after this, the machine can find out the correct meaning of the word in the particular context and can translate the meaning of that word into the correct word in another language. The process of finding the correct meaning of the polysemy word using machine by analyzing the context in which the polysemy word has been used is referred as Word Sense Disambiguation (WSD).

Although different methods have been tested to find the correct sense of the polysemy word, accuracy at satisfactory level has not been obtained yet. Among the different methods used for WSD, this research focused its study on the knowledge-based approach. The knowledge-based approaches use the resources such as dictionaries thesauri, ontology, collocation etc to disambiguate a word in a given context [1].

These days, the WordNet is becoming popular as a resource to be used in knowledge-based approach to disambiguate the meanings of polysemy words. WordNet is a lexical database developed at Princeton University for English language [2]. The WordNet organizes nouns, verbs, adjectives and adverbs into the groups of synonyms and describes the relationships between these synonym groups forming a semantic network among the words. After the development of English WordNet, many other WordNet on other languages such as Spanish WordNet, Italian WordNet, Hindi WordNet etc were built. These WordNet are used as one important resource to disambiguate the different meanings of a polysemy words in respective languages. To disambiguate the meaning of a polysemy word using the WordNet, the related words from synset, gloss and different levels of hypernym are collected from the WordNet database and these related words are compared to find the overlaps using different WSD algorithm. However, the collection of related words from synset, gloss and different level of hypernym that are taken from the WordNet contains only very few words that can be used to disambiguate the correct sense of a polysemy word in the given context [3]. This increases only the computational overhead for the WSD algorithm and for the system. Moreover, according to [3] as we go increasing the levels of hypernym to collect the related words, the sense of the words becomes more general and the two different senses of a polysemy word are found to have the same hypernym. This therefore creates another ambiguity in ambiguity. To overcome this problem, we have developed a new model to organize both the single sense and multi-sense words.

## 2. RELATED TASKS

In 1986, Lesk Michael [4] developed an algorithm called Lesk algorithm to identify senses of polysemy words. He used the overlap of word definition from the *Oxford Advanced Learner's Dictionary of Current English* (OALD) to disambiguate the word senses.

Banerjee and Pedersen [5] adapted the original Lesk algorithm to use the lexical database WordNet. They used Senseval-2 word sense disambiguation exercise to evaluate their system and the overall accuracy was found to be 32%. To compare two glosses, they used the longest sequence of one or more consecutive words that occurs in both glosses. For each overlap, a square of the number of words in the overlap is calculated and the final score is the sum of all overlaps.

Sinha M. et. al. [6] developed automatic WSD for Hindi language using Hindi WordNet at IIT Bombay. They used statistical method for determining the senses. The system could disambiguate the nouns only. They compared the context of the polysemy word in a sentence with the contexts constructed from the WordNet. They evaluated the system using the Hindi corpora provided by the Central Institute of Indian Languages (CIIL). The accuracy of their algorithm was found in the range from 40% to 70%. They used simple overlap method to determine the winner sense.

In [7], Shrestha N. et. al. used the Lesk algorithm to disambiguate the Nepali ambiguous words. They modified the Lesk algorithm in such a way that only the words in the sentence without their synset, gloss, examples and hypernym are taken as context words. Each word in the context words is compared with each word in the collection of words formed by the synset, gloss, examples and hypernym of each sense of the target word. They did not include the synset, gloss, example and hypernym of the context words in the collection of context words. Moreover, the number of example for each sense of the target word was only one.

In [3], Dhungana and Shakya used the adapted Lesk algorithm to disambiguate the polysemy word in Nepali language. The experiments performed on 348 words (including the different senses of 59 polysemy words and context words) with the test data containing 201 Nepali sentences shows the accuracy of their system to be 88.05%. This accuracy is found to be increased by 16.41%, in compared to the accuracy of the system developed by [7].

Context word refers to the word that is used in context with the target word. The target word is the word which has multiple senses at different contexts and its meaning needs to be disambiguated in the given context. Hypernymy is a semantic relation between two synsets to show super-set hood (is a kind of) relation. For example, बेलपत्र (Transliteration: Belpatra) is a kind of पात (Transliteration: Paat, Transliteration in English: Leaf). Therefore, पात is a hypernymy and बेलपत्र is hyponymy [8].

## 3. STATEMENT OF RESEARCH PROBLEM

### 3.1. Problem Statement

From analyzing many research works on WSD using WordNet such as [4], [5], [6], [7], [9] and [3], it was noticed that the WordNet is very useful resource to use for word sense disambiguation. However, it is not exactly suitable for knowledge-based, overlap selection WSD approaches. The reason is that the WordNet is built for general purpose in NLP tasks but is not focused for WSD. WordNet contains huge amount of information for words that are arranged with semantic relation. In all WSD methods, the WordNet is used to take a large number of words to disambiguate the meaning of multi-sense word. However, only very few words taken from the WordNet are used to disambiguate the different senses of a multi-sense word. In this sense, all the efforts such as processing time for CPU and memory to store large number of unused words are wasted. Furthermore, it is noticed that the words taken from the WordNet to disambiguate the multiple meaning of the multi-sense word itself creates the ambiguity resulting in decrease in accuracy.

Another important point noticed from [3] is that when deeper levels of the hypernymy from the WordNet are used, the correctly disambiguated polysemy words are also incorrectly disambiguated. The reason is that the deeper (probably from second and/or third) levels of Hypernymy of all the different sense of a polysemy word are same as shown in Fig.1. That's why, there is no any basis to disambiguate the different sense of the polysemy word.

*Example: WSD and Hypernymy in English WordNet:* Let us consider a polysemy word Pen in English language. In WordNet, the word has five different meanings as a noun. The five meanings of the word Pen with the different levels of hypernymy are shown in Fig. 1. First look at the Fig.1b and Fig.1c. Here, we can see that the only the meaning of two different words are different. The hypernymy of these two senses are the same. In such case, there is no meaning of using the hypernymy of the two different senses of the polysemy word to disambiguate their meanings. Inclusion of these hypernymy is just waste of computational effort and waste of memory for the system.

### 3.2. Solution Approach

What we need is that the organization of the polysemy words must be organized in such a way that it should contain only those words that are sufficient to disambiguate the different senses of a polysemy word and it should not again introduce ambiguity as it is in the case of using WordNet. Although WordNet is very useful lexical resource that can be used to disambiguate the different meanings of a polysemy word, it has still some limitations to use in WSD algorithms as discussed in previous section. Therefore, to improve the accuracy of the knowledge-based overlap selection WSD algorithms dramatically, we need to develop a new logical model that organizes the words in such a way that it disambiguates the meanings of a polysemy word correctly and efficiently resulting in higher accuracy than that can be obtained from the use of WordNet.

### 3.2.1. Organization of Clue Words

Unlike in WordNet, we organized the different senses of polysemy word as well as single sense words. We grouped each sense of a polysemy word based on the verb, noun, adverb and adjective with which the sense of the polysemy word can be used in a sentence. The organization of the single sense is also done in the same way. This organization of words results in a new model of WordNet which is focused on WSD. This new model of WordNet contains all the necessary and sufficient words/information that can be used to disambiguate the senses of a polysemy word. It restricts the unnecessary words/information and therefore does not create ambiguity in ambiguity like WordNet. For instance, a sample organization of different meanings of a polysemy word Pen in English language as a noun is shown in Fig. 2.

| Polysemy Word | (PoS) Senses | Used with verb | Used with noun | ... | Used with Preposition |
|---|---|---|---|---|---|
| Pen | (n) a writing implement with a point from which ink flows | Write, draw ... | Copy, Paper ... | ... | With, by ... |
| | (n) an enclosure for confining livestock | Be, keep... | Rabbit, Dog... | ... | In, Inside... |
| | ... | ... | ... | ... | ... |

Figure 2. Organization of different senses of noun *Pen* in new model of WordNet based on clue words

### 3.2.1. Algorithm

The algorithm is very simple. First, we find collection of clue words for each sense of a polysemy word. The number of collection of words for a polysemy word is equal to the number of different senses of the polysemy. Let say these collection of words be T1, T2, T3,...,Tn where n is the number of different senses that the polysemy word has. After this, we find the collection of words (say C) from the context window.

## 4. INTENDS

The main objectives of this research work include:

1. To organize the senses of polysemy words based on clue words.

2. To develop a new model of WordNet based on clue words for both polysemy and single sense words.

3. To test whether using the conventional or using the new model WordNet, the WSD algorithm gives the higher accuracy.

## 5. HYPOTHESIS

The conventional WordNet is very useful resource for language learning and language processing tasks. However, it is not suitable for knowledge-based WSD algorithms to get higher accuracy. We belief that organizing the different senses of polysemy words and context words based on clue words best suits for knowledge-based WSD algorithms resulting in higher accuracy. Moreover this reduces the computational effort for the system and saves the system memory while processing.

## 6. RESEARCH METHODOLOGY

Depending upon the nature of our research, we used experiment type of research strategy. We used the algorithm and test data already developed by [3] to test our new model of WordNet for WSD. Based on the research question, we set up two experiments with different settings. In first experiment, we used the system developed by [3] using the conventional sample Nepali WordNet. In second experiment, we only replaced the conventional sample Nepali WordNet by our new model of WordNet organized with clue words keeping all the other settings constant as in first experiment.

### 6.1. Experiment 1

**Lexical Resource**: In first experiment, we used the conventional sample Nepali WordNet developed by [3]. This conventional sample Nepali WordNet contains 348 words including 59 polysemy words. The words in this WordNet are organized as in English WordNet.

**Experiment Setting**: In this setting, we included the synset, gloss, example and hypernym of each word of in the collection of context words to form the final collection of context words and compared these words with the collection words of each sense of target word to determine overlaps. The number of examples for each word provided in this case is more than four in average. Moreover, we did not include the target word in the collection of context word.

### 6.2. Experiment 2

**Lexical Resource**: In this Second experiment, we used our new model of WordNet that is focused for word sense disambiguation.

**Experiment Setting**: The experiment setting is same as the setting in Experiment 1 except that we replaced the conventional sample Nepali WordNet by our new model of WordNet organized with clue words.

## 7. RESULT AND DISCUSSIONS

After setting the two experiments, we executed the experiments using the test data developed by [3]. The test data consists of 201 different Nepali sentences each containing polysemy word. In experiment 1, we found that the polysemy words in 117 test sentences out of 201 are correctly disambiguated. This shows the accuracy of the system with conventional sample Nepali WordNet found to be 88.059%. In experiment 2, we found that the polysemy words in 184 test sentences out of 201 are correctly disambiguated. At this time, the accuracy of the system is found to be 91.543%. Therefore, the accuracy of the system using our new model of WordNet that is specific to WSD is found to be increased by 3.484%. The accuracies obtained in the two experiments are shown in Table 1.

Table 1. Heading and text fonts.

| Experiment No | No. Of polysemy words that are correctly disambiguated | No. Of polysemy words that are correctly disambiguated | Accuracy in percentage |
|---|---|---|---|
| 1 | 177 | 24 | 88.059 |
| 2 | 184 | 17 | 91.543 |

## 8. CONCLUSIONS

The WordNet organizes the words in the lexical database based on their meanings of the words instead of their forms as in dictionaries. It groups the nouns, verbs, adjectives and adverbs together into synonym sets, each expressing a distinct concept [2]. The words in a synonym set can be interchangeably used in many contexts. The main relationship among the words in WordNet is the synonym. From analyzing many research works on WSD using WordNet such as [4], [5], [6], [7], [9] and [3], we have noticed that the WordNet is not exactly suitable to use with knowledge-based, overlap selection WSD approaches. The reason is that the WordNet is built for general purpose in NLP tasks but not focused for WSD. In all WSD methods that used WordNet, the WordNet is used to take a large number of words to disambiguate the meaning of multi-sense word. But it is noticed that only very few words taken from the WordNet are used to disambiguate the different senses of a multi-sense word. In this sense, all the efforts such as processing time for CPU and memory to store large number of unused words are wasted. To defeat with these problems of WordNet, we developed a new model of WordNet that organizes the different senses of polysemy words as well as single sense word using the clue words. The experiment shows that the accuracy of the system using our new model of WordNet is found to be increased by 3.484% in compare with the accuracy of the system in [3]. This evidence leads us to conclude that our new model of WordNet gives the higher accuracy for the knowledge-based WSD algorithms rather than using the conventional WordNet.

## ACKNOWLEDGEMENTS

We would like to kindly express our sincere thanks to Madan Puraskar Pustakalaya (MPP), Patan Dhoka, Lalitpur for providing Nepali corpus and Nepali online dictionary which formed the main resources of this research. We are equally thankful to Mr. Balram Prasain, Lecturer, Tribhuwan University for his valuable help in this research.